\documentclass[12pt]{article}

\usepackage{geometry}
\usepackage{newtxtext,newtxmath}

\usepackage{graphicx}
\usepackage{amsmath}
\usepackage{booktabs}
\usepackage{hyperref}
\usepackage{enumitem}
\usepackage{tikz}
\usepackage{pgfplots}
\usepackage{fancyhdr}

\pgfplotsset{compat=1.18}

\usepackage{hyperref}
\usepackage{cleveref}

\begin{document}

\begin{center}
\textbf{\fontsize{15}{18}\selectfont Mitigating Early Training Collapse in CTR Models}
\end{center}

\vspace{0.5em}

\noindent
\textbf{Ergun Biçici$^{1}$, Erkan Çetinyamaç$^{1}$}

\noindent
$^{1}$ Intelligent Application Development, Huawei Türkiye R\&D Center, Istanbul, Turkey

\vspace{0.5em}

\noindent
\textbf{*Corresponding author:}  
Ergun Biçici, Huawei Türkiye R\&D Center, Istanbul, Turkey.  
Email: ergun.bicici@huawei.com

\vspace{1em}

\section*{Abstract}
Deep neural models for click-through rate prediction often exhibit a sharp decline in validation performance immediately after the first training epoch despite continued improvement in training loss. This instability restricts effective learning and limits model performance. In this study, we analyze this behavior using large-scale industrial datasets and evaluate practical mitigation strategies. While reducing the learning rate provides only incremental gains, controlling feature sparsity yields substantial improvements. Removing highly sparse features and aggregating infrequent feature values stabilizes training, extends useful learning beyond a single epoch, and improves both offline evaluation metrics and online system performance.

\section*{Keywords}
CTR prediction, overfitting, sparsity, embeddings, recommender systems

\section*{Abbreviations}
CTR: Click-Through Rate; CVR: Conversion Rate; AUC: Area Under Curve; PRAUC: Precision-Recall AUC

\section{Introduction}
Deep learning models are widely used in CTR prediction due to their ability to capture complex feature interactions \cite{FINT,FFCP,Masknet}. However, in industrial environments with high-cardinality categorical inputs, models frequently reach peak validation performance after a single epoch and degrade thereafter. 
This effect has been observed in prior work and is associated with rapid overfitting in sparse feature spaces \cite{OneEpoch}.

\begin{figure}[t]
\centering
\begin{tikzpicture}
\begin{axis}[
width=0.9\linewidth,
height=6cm,
xlabel={Epoch},
ylabel={Performance (AUC)},
xmin=0.8, xmax=5.2,
ymin=0.55, ymax=1.02,
legend style={at={(0.98,0.98)},anchor=north east},
grid=major,
tick align=outside,
]

\addplot[
blue,
thick,
mark=o
]
coordinates {
(1,0.75)
(2,0.85)
(3,0.92)
(4,0.96)
(5,0.98)
};

\addplot[
red,
thick,
mark=square
]
coordinates {
(1,0.80)
(2,0.70)
(3,0.68)
(4,0.66)
(5,0.65)
};

\addplot[
black,
dashed,
thick
] coordinates {(1,0.55) (1,1.02)};

\node[anchor=south west] at (axis cs:1.05,0.95)
{\small One-Epoch Overfitting};

\legend{training, validation}

\end{axis}
\end{tikzpicture}
\caption{Illustration of the one-epoch phenomenon within a U-shaped training regime. Training performance continues to improve while validation performance peaks after the first epoch due to overfitting.}
\label{OneEpochTraining}
\end{figure}
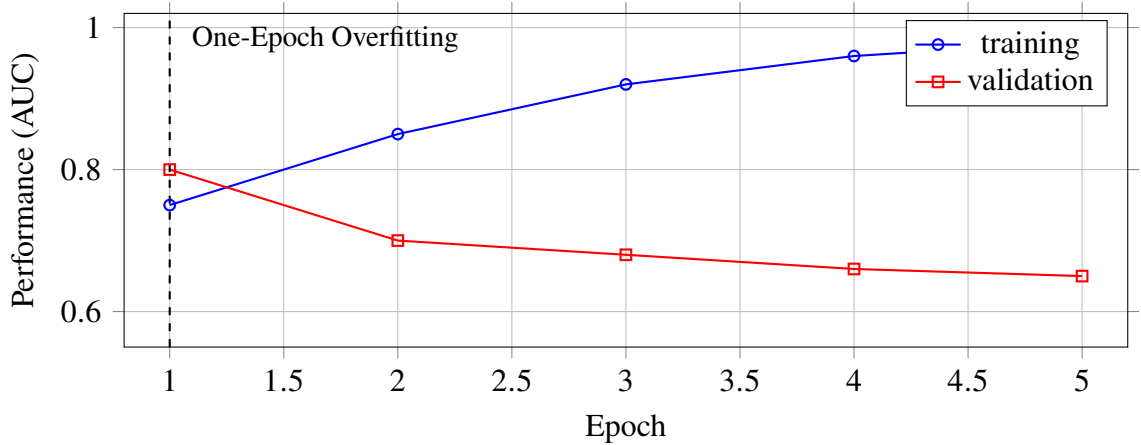

CTR datasets exhibit long-tailed distributions where a small subset of feature values dominates frequency while most occur rarely \cite{PICASSO2022}. This imbalance increases variance and encourages rapid memorization. Unlike vision tasks where overfitting progresses gradually \cite{ResNet}, CTR models often degrade abruptly.
\Cref{OneEpochTraining} depicts one-epoch phenomenon within a U-shaped training regime~\cite{DoubleDescent2019}.

\section{Materials and Methods}

\subsection{Problem Mechanism}
This behavior arises from the interaction of model capacity, optimization dynamics, and data sparsity. Embedding layers assign parameters to each categorical value, but rare values receive very few updates, leading to unstable representations. Adaptive optimizers accelerate convergence \cite{kingma2015adam}, allowing the model to quickly fit noise. As a result, the model memorizes infrequent patterns early, causing validation performance to deteriorate.

\subsection{Proposed Strategies}
We evaluate three approaches:

\begin{itemize}[leftmargin=*]
    \item \textbf{Learning rate reduction:} Slows convergence but does not eliminate early overfitting.
    \item \textbf{Sparse feature removal:} Eliminates high-cardinality features with low signal.
    \item \textbf{Value filtering:} Retains only frequent values, mapping others to a shared token.
\end{itemize}

\section{Results and Discussion}

Experiments use industrial-scale CTR datasets with hundreds of millions of samples and strong class imbalance. Performance is evaluated using AUC, log-loss, and PRAUC.

\subsection{Offline Results}

\begin{table}[h]
\centering
\caption{Offline CTR performance summary (relative to baseline).}
\label{tab:results}
\begin{tabular}{lcccc}
\toprule
Method & AUC & Logloss & PRAUC & Epochs \\
\midrule
Baseline & 0.8445 & 0.3687 & 0.2797 & 2 \\ \hline
Lower LR & +0.15\% & -4.1\% & +1.2\% & 2 \\
Feature Removal & +0.33\% & -5.5\% & +1.5\% & 3--4 \\
Value Filtering & +0.04\% & -0.3\% & +0.6\% & 2--3 \\
\bottomrule
\end{tabular}
\end{table}

The results in~\Cref{tab:results} show that reducing the learning rate yields small improvements but does not prevent early degradation. Removing sparse features provides the largest gains and extends stable training to multiple epochs. Value filtering further improves robustness by reducing variance in embedding updates.

\subsection{Online Results}

Deployment in an online advertising system shows consistent improvements in key metrics, including conversion rate and revenue efficiency, confirming that offline gains transfer to production. The results in~\Cref{OnlineResults} show that total advertiser value (TAV)~\cite{Meta2025}, which is the main improvement metric, increased reached $1.88\%$. 

\begin{table}[h]
\centering
\caption{Online performance improvements}
\begin{tabular}{cccc}
\toprule
TAV & RPM & eCPM & CVR \\
\midrule
1.88\% & 2.02\% & 1.1\% & 3.39\% \\
\bottomrule
\end{tabular}
  \label{OnlineResults}
\end{table}

The improvements confirm that sparsity reduction translates to real-world gains.

\section{Conclusion}

The results indicate that early training collapse in CTR models is primarily driven by feature sparsity rather than optimization dynamics alone. High-cardinality inputs introduce many weakly supported parameters, enabling rapid memorization. Reducing feature space complexity improves generalization and enables stable multi-epoch training. While optimization adjustments offer limited benefit, controlling feature sparsity is highly effective. Removing sparse features and aggregating rare values improves both stability and performance. The proposed methods are simple, effective, and compatible with existing systems.

\section*{Acknowledgements}
None.

\section*{Conflict of Interest}
The authors declare no conflict of interest.

\bibliographystyle{plain}

\end{document}